\documentclass[11pt]{article}

\usepackage[preprint]{acl}

\usepackage{times}
\usepackage{latexsym}
\usepackage{amsfonts}
\usepackage{enumitem}
\usepackage{amsmath}
\usepackage{booktabs}
\usepackage{multirow}
\usepackage{makecell}
\usepackage{tcolorbox}
\tcbuselibrary{breakable}
\usepackage[T1]{fontenc}

\usepackage[utf8]{inputenc}
\usepackage{tcolorbox}
\tcbuselibrary{listings}

\usepackage{microtype}
\usepackage{booktabs}
\usepackage{colortbl}
\usepackage{soul}
\usepackage{xcolor}

\usepackage{inconsolata}
\usepackage{ulem}
\usepackage{graphicx}
\usepackage[table]{xcolor}

\definecolor{codeblue}{RGB}{0,0,139}
\definecolor{codegreen}{RGB}{0,100,0}
\definecolor{codered}{RGB}{178,34,34}

\usepackage{tabularx}
\usepackage{booktabs}
\usepackage{multirow}
\usepackage{makecell}

\usepackage{tcolorbox}
\usepackage{listings}
\usepackage{xcolor}
\usepackage[HTML]{xcolor}

\usepackage{longtable}
\usepackage{graphicx} 
\usepackage{xspace}   
\usepackage{fontawesome}   
\newcommand{\fleaf}{\textsuperscript{\fontsize{5.5pt}{6pt}\selectfont \faLeaf}}

\newcommand{\email}[1]{\href{mailto:#1}{\texttt{#1}}}

\title{DocDancer: Towards Agentic Document-Grounded Information Seeking}

\author{
Qintong Zhang\textsuperscript{$\heartsuit$}\thanks{Equal Contributions. $\fleaf$Jialong Wu is the project leader. },
Xinjie Lv\textsuperscript{$\heartsuit$}$^{*}$,
Jialong Wu\textsuperscript{$\heartsuit$}$^{*}$$\fleaf$,
Baixuan Li$^{*}$,
Zhengwei Tao\textsuperscript{$\heartsuit$}, \\
\textbf{
Guochen Yan\textsuperscript{$\heartsuit$},
Huanyao Zhang\textsuperscript{$\heartsuit$},
Bin Wang\textsuperscript{$\diamondsuit$},
Jiahao Xu\textsuperscript{$\clubsuit$},
Haitao Mi\textsuperscript{$\clubsuit$},
Wentao Zhang\textsuperscript{$\heartsuit$}\thanks{Corresponding Author.}
} \\
\textsuperscript{$\heartsuit$}Peking University,
\textsuperscript{$\diamondsuit$}Shanghai AI Lab,
\textsuperscript{$\clubsuit$}Tencent AI Lab
\\
\email{wujialongml@gmail.com}, \email{wentao.zhang@pku.edu.cn} \\
}

\begin{document}
\maketitle
\begin{abstract}
Document Question Answering (DocQA) focuses on answering questions grounded in given documents, yet existing DocQA agents lack effective tool utilization and largely rely on closed-source models.
In this work, we introduce \textbf{DocDancer}, an end-to-end trained open-source Doc agent.
We formulate DocQA as an information-seeking problem and propose a tool-driven agent framework that explicitly models document exploration and comprehension.
To enable end-to-end training of such agents, we introduce an \textit{Exploration-then-Synthesis} data synthesis pipeline that addresses the scarcity of high-quality training data for DocQA.
Training on the synthesized data, the trained models on two long-context document understanding benchmarks, MMLongBench-Doc and DocBench, show their effectiveness.
Further analysis provides valuable insights for the agentic tool design and synthetic data.
\end{abstract}

\section{Introduction}
Understanding and answering questions over long, multi-modal documents is a critical capability for real-world intelligent systems~\cite{tkaczyk2015cermine,liu2025comprehensive}.
Document Question Answering (DocQA) lies at the core of document-centric intelligence, enabling models to access, reason over, and synthesize information from complex and heterogeneous document sources.

\begin{figure}[t]
  \centering
  \includegraphics[width=\linewidth]{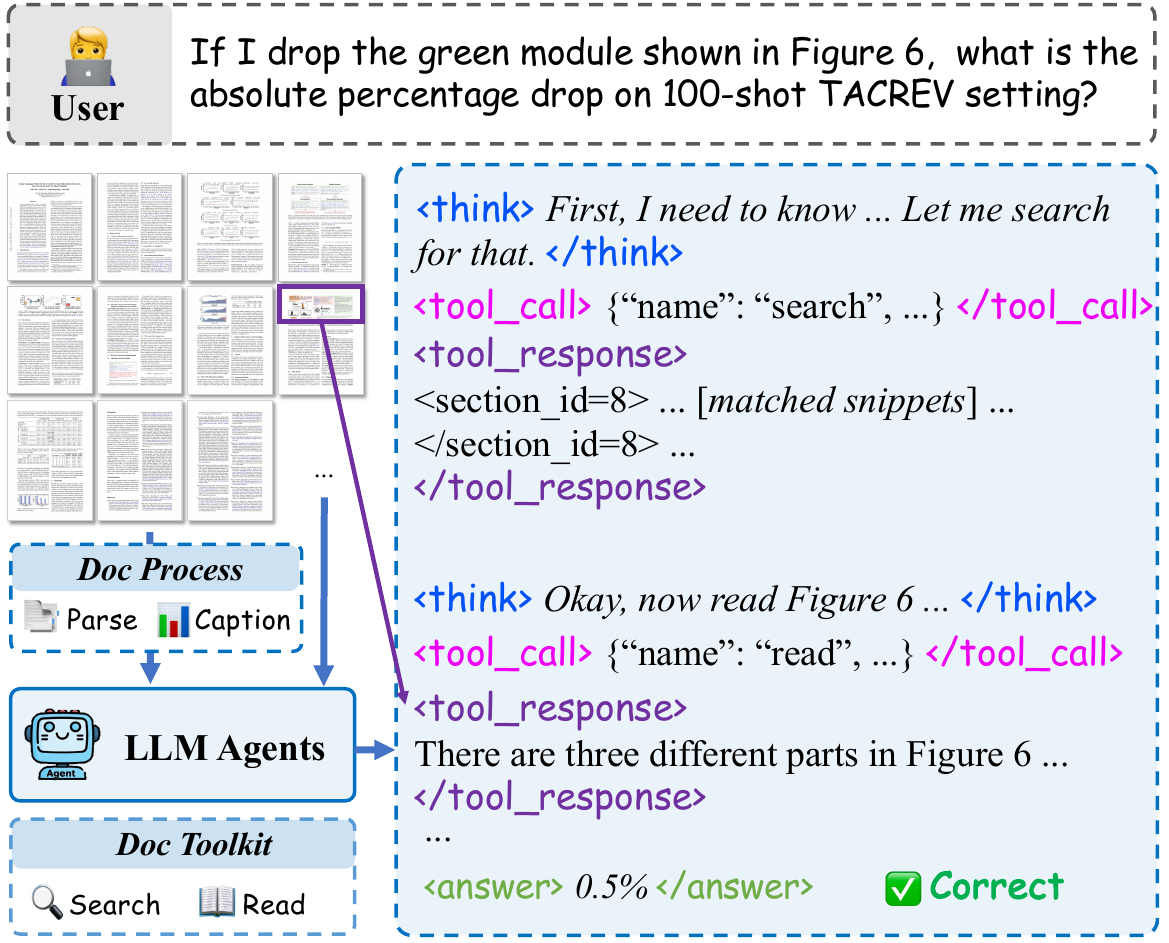}
   \caption{The overall of \textbf{DocDancer} for document-grounded information seeking, where \textit{search} and \textit{read} tools for effective document retrieval and comprehension over processed documents.}
   \label{fig:workflow}
\end{figure}

Existing DocQA methods can be broadly categorized into three paradigms.
The first paradigm relies on optical character recognition (OCR) to convert documents into plain text, which is then processed by downstream language models~\cite{xu2020layoutlm}.
The second paradigm adopts embedding-based retrieval mechanisms, most commonly instantiated through retrieval-augmented generation (RAG), to identify and incorporate relevant document segments during inference~\cite{saad-falcon-etal-2024-pdftriage}.
More recently, agent-based paradigms have gained increasing attention, as they better support complex scenarios that require iterative exploration, tool invocation, and multi-step reasoning over long and structured documents~\cite{sun-etal-2025-docagent,zhu2025doclens}.
Recent advances in large language models (LLMs) \cite{qwen3technicalreport, liu2025deepseek} enable such agents to dynamically decompose queries, interact with documents, and adapt to intermediate observations, alleviating the limitations of OCR- and RAG-based approaches.
Despite their promise, existing DocQA agents are typically implemented as prompt-based pipelines, with limited learning of autonomous agentic behaviors.

In contrast, we aim to train the \textbf{first} end-to-end DocQA agent model that is explicitly grounded in information-seeking principles, moving beyond prompt-based agent designs.
We first formulate DocQA as an agentic information-seeking problem and design a tool-centric agent framework that decomposes document understanding into two complementary capabilities.
Specifically, we introduce efficient \texttt{search} tools for global information acquisition and fine-grained \texttt{read} tools for localized comprehension.
This design enables the agent to actively explore long documents, iteratively refine its hypotheses, and dynamically adapt its strategy based on intermediate observations.
Notably, when instantiated with a proprietary LLM, our framework achieves state-of-the-art performance and exceeds reported human-level performance.

Furthermore, a key bottleneck in training such agent models is the scarcity of high-quality DocQA pairs~\cite{huang2025airqa}, as most publicly available datasets provide only test splits and lack sufficiently annotated training data.
To address this challenge, we propose an \textit{Exploration-then-Synthesis} DocQA generation pipeline that progressively enhances QA pairs from easy to hard.
Specifically, we first explore a source document through intent-guided, tool-augmented interactions to collect grounded evidence (the \textit{Exploration} stage), and then synthesizes high-quality document-grounded QA pairs via multi-observation reasoning (the \textit{Synthesis} stage).
We then train our DocQA agent, \textbf{DocDancer}, on the synthesized dataset, instantiating it with two open-source backbones, Qwen3-4B-Thinking-2507 and Qwen3-30B-A3B-Thinking-2507~\cite{qwen3technicalreport}.
Despite being trained with \textbf{only 5,000} instances, both variants achieve competitive performance, with the 30B-A3B model attaining state-of-the-art results in several settings.

Extensive experiments are conducted on two long-context document understanding benchmarks, MMLongBench-Doc~\cite{ma2024mmlongbench} and DocBench~\cite{zou2025docbench}.
The results demonstrate the effectiveness of the proposed \textbf{DocDancer}.
Further analyses provide insights into document parsing strategies, tool design, and the role of synthetic data in agent learning. In summary, our contributions are three-fold:
\begin{itemize}[itemsep=0.5pt, parsep=0pt]
\vspace{-2mm}
    \item \textbf{Effective Agentic DocQA Framework}:
    We propose a tool-driven DocQA agent framework grounded in information-seeking principles, which achieves SOTA performance when paired with a proprietary LLM.
    \item \textbf{Autonomous Data Synthesis Pipeline}:
    We introduce an \textit{Exploration-then-Refine} data synthesis pipeline that generates high-quality training data for learning agentic behaviors.
    \item \textbf{Empirical Performance}: Our method achieves state-of-the-art results and provides practical insights into effective and efficient agentic system design.
\end{itemize}
\section{Related Work}

\textbf{Document Question Answering Methods.}
Traditional DocQA methods rely on OCR-based pipelines~\cite{ding2022v} or end-to-end vision–language models~\cite{sukh2025ocr, hu2025mplug}, but both are constrained by limited input length and struggle with long documents~\cite{ma2024mmlongbench,zou2025docbench,dong2025mmdocir}.
Retrieval-augmented generation~\cite{zhang2024cream, dong2025mmdocir, dong2025benchmarking} improves scalability, yet most approaches decouple retrieval and reasoning in a single-shot manner, making them brittle to retrieval errors and ineffective for complex, multi-step queries~\cite{zhang2025ocr}. 
Recent agent-based DocQA systems~\cite{wu2025doc,sun-etal-2025-docagent,dong2025doc} address these issues through iterative document navigation and reading, but they predominantly depend on prompt-engineered, closed-source LLMs.
In this work, we aim to train an open-source document agent with learnable behaviors for robust and scalable DocQA.
 
\noindent \textbf{Synthetic Data for Agent Training.}
High-quality training data is critical for training agents.  
Due to its scalability, rapid iteration, and inherent trainability, synthetic data offers significant advantages over manually annotated data, serving as a highly effective alternative to human-labeled datasets for agent learning~\cite{liu2025deepseek,team2025tongyi}.
Prior work has demonstrated that large-scale agent-synthesized data can be effectively generated for search agents~\cite{wu2025webdancer,li2025websailor,tao2025webshaper}, code agents~\cite{yang2025swe}, GUI agents~\cite{sun2025genesis,guo2025auto} and general-purpose agents~\cite{fang2025towards,prabhakar2025apigen}. 
In contrast, this work focuses on the DocQA agent setting.
Existing DocQA datasets are primarily constructed through semi-automated~\cite{van2023document,dong2025benchmarking} or expert-annotated~\cite{hendrycks2103cuad,deng2025longdocurl} processes, both of which require substantial human involvement or result in questions that lack sufficient depth.
Inspired by advances in search agents, we formulate DocQA as an agentic information-seeking problem, with the goal of synthesizing high-quality training data tailored for DocQA agents.

\section{Methods}
\subsection{Agent Setup}

\noindent \textbf{Framework.} 
We adopt the vanilla ReAct~\citep{yao2022react} as the agent's framework, which synergizes reasoning and acting. 
In this paradigm, the agent generates both a reasoning trace (thought), $\tau$, and a subsequent action, $a$, in an interleaved manner. 
This process forms a trajectory, $\mathcal{H}_T$, which is a sequence of thought-action-observation triplets:
\begin{equation}
    \mathcal{H}_T=(\tau_0,a_0,o_0,\dots, \tau_i,a_i,o_i, \dots,\tau_{T},a_{T}), 
\end{equation}
where $a_T$ represents the final answer to the given task.
At any given step $t\leq T$, the agent's policy, $\pi$, generates the current thought $\tau_t$ and action $a_t$ based on the history of all previous interactions, $\mathcal{H}_{t-1}$:
\begin{equation}
\tau_t, a_t \sim \pi(\cdot | \mathcal{H}_{t-1}).
\end{equation}

\begin{figure*}[t]
  \centering
   \includegraphics[width=\linewidth]{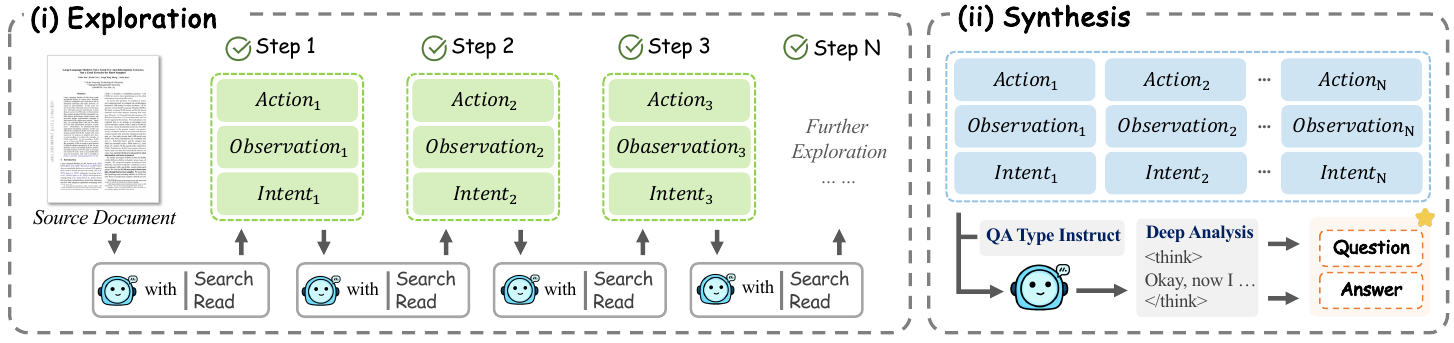}
   \caption{\textbf{Overall of the \textit{Exploration-then-Synthesis} framework}. 
   (i) \textit{Exploration} stage iteratively interacts with the source document through Action($u$)–Observation($y$)–Intent($i$) steps.
    (ii) \textit{Synthesis} stage aggregates the collected evidence to generate the final question and answer.
    We present a concrete case illustrating the whole generation process in Appendix~\ref{app:case_study}.}
   \label{fig:data_synthesis}
\end{figure*}

Inspired by \textit{The Bitter Lesson}~\cite{sutton2019bitter}, we employ a single-agent setup with carefully selected, highly effective tools, rather than relying on multi-agent designs or test-time scaling.

\noindent \textbf{Document Processing.}
Prior works~\cite{sun-etal-2025-docagent} show that an XML-based hierarchical representation for document outlines that organizes parsed content into nested trees, using sections as partitioning units and elements such as text, images, and tables as nodes. While this structure enables efficient positioning and search, it suffers from structural and content inaccuracies and does not incorporate retrieval-aware visual information, which limits its applicability to agent-based processing of long, visually rich documents.
To address these issues, we substantially enhance the document outline.
For content accuracy, we leverage MinerU2.5~\cite{niu2025mineru2} for high-precision layout analysis and extraction, defining 17 element types and enriching outline nodes with layout and semantic attributes while removing structurally irrelevant elements such as headers and footers. For structural accuracy, title elements are visually cropped and clustered to infer hierarchical levels, enabling fine-grained section segmentation and reducing information loss in long documents. To improve visual retrieval, we generate captions for images and charts using an multimodal model $M_m$ and incorporate them as auxiliary information, allowing the outline to better align and retrieve visual content.

\noindent \textbf{Tool Design.} We point out that DocQA can be naturally formulated as an \textit{agentic information-seeking} task in which the external information source is restricted to the given documents. Accordingly, our tool design aims to enable agents to efficiently and effectively locate and extract relevant information from documents, while keeping the overall toolkit complexity low to ensure ease of use for agent models. Specifically, we design the following two tools for DocDancer:

\begin{itemize}[leftmargin=1.5em, itemsep=2pt, topsep=1pt, parsep=0pt]
    \item \textbf{\textit{Search.}} Conducts keyword-based full-text search over the given documents, returning the section IDs, page numbers, and surrounding text snippets for each match. A visible window is used to constrain the snippet length for efficient localization. This tool provides the agent with \underline{global} textual signals for guiding subsequent information access.
    \item \textbf{\textit{Read.}} Given a goal and a set of section IDs, the tool performs \underline{fine-grained} reading to extract goal-relevant information from the specified sections. This includes (i) local textual information, consisting of all text within the section; (ii) local visual information, consisting of images and tables within the section, together with a page-level screenshot that captures the full layout of the page containing the section. Subsequently, a multimodal summarization model $M_m$ is used as an auxiliary reader to jointly integrate textual and visual inputs and return consolidated goal-relevant content.
\end{itemize}

This design deliberately integrates textual and visual signals, capturing both localized evidence and global layout cues, while keeping the toolkit limited to two tools to facilitate efficient utilization.

\subsection{Data Synthesis}
\label{sec:data}
It is crucial to curate complex and diverse Document DocQA pairs that are capable of eliciting multi-step reasoning, goal decomposition, and rich interaction trajectories. 
To this end, we first construct a broad and heterogeneous collection of PDF documents to serve as the grounding corpus for question answering. 
We then synthesize QA pairs based on these documents, ensuring coverage of diverse reasoning patterns and document structures.

\begin{figure}[!htbp]
  \centering
   \includegraphics[width=0.98\linewidth]{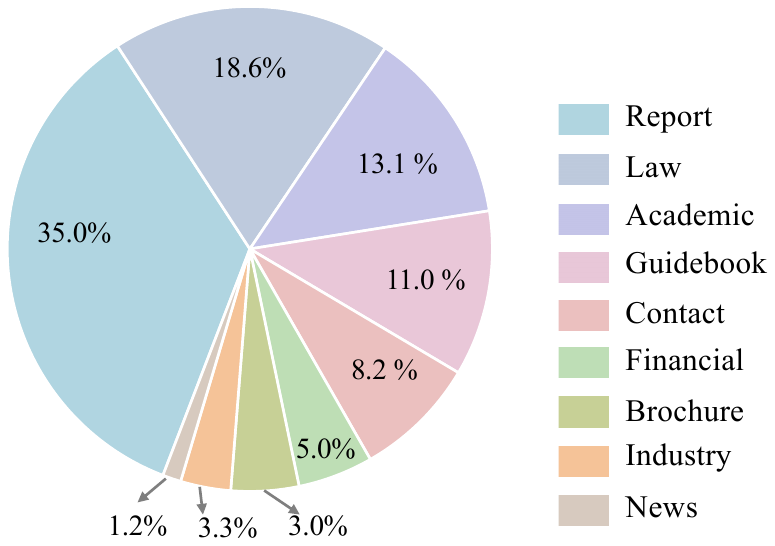}
   \caption{Distribution of document used to synthesise.}
   \label{fig:PDF_distribution}
\end{figure}

\noindent \textbf{Sources.}
To construct a robust and diverse dataset for document-based question answering, we select four representative datasets, LongDocURL~\cite{deng2025longdocurl}, MMDocRAG~\cite{dong2025benchmarking}, CUAD~\cite{hendrycks2103cuad} and DUDE~\cite{van2023document}, that cover long-context understanding, multimodal retrieval, legal expertise, and complex layout analysis. 
These sources provide the foundational PDF documents used for our automated QA generation pipeline.
The distribution of the collected PDF documents is illustrated in Figure~\ref{fig:PDF_distribution}.

\noindent \textbf{Exploration-then-Synthesis Framework.} 
We propose a two-stage framework for DocQA generation, consisting of an \textit{Exploration} Stage and a \textit{Synthesis} Stage as shown in Figure~\ref{fig:data_synthesis}.
The overall objective is to transform a source document into a diverse and high-quality set of grounded QA pairs through iterative interaction and reasoning.

\textbf{\textit{Exploration} Stage.} Given a source document $\mathcal{D}$, utilze an LLM $M_e$ to iteratively interact with 
$\mathcal{D}$ and collect information relevant to potential QA pairs.
Conditioned on the interaction history $h_t$ and the document $\mathcal{D}$, we employ model $M_s$ jointly generates an intent-action pair $(i_t, a_t)$:
\begin{equation}
(i_t, u_t) \sim \pi_{M_e}(i, u \mid h_t, \mathcal{D}),
\end{equation}
where $i_t$ denotes the exploration intent and $u_t \in \mathcal{A}$ corresponds to invoking a document-grounded tool such as \textit{Search} or \textit{Read}, which is the same as the agent’s tool action.
The construction of a question implicitly induces the strategy required to resolve it.
The explicit modeling of intent helps prevent uninformative exploration, guiding the agent toward more concrete, goal-directed trajectories~\cite{pahuja2025explorer}.
Executing action $a_t$ yields an observation:
\begin{equation}
y_t = \mathcal{T}(a_t, \mathcal{D}),
\end{equation}
where $\mathcal{T}$ denotes the document interaction interface. The interaction history is then updated as:
\begin{equation}
h_{t+1} = h_t \cup \{(i_t, u_t, y_t)\},
\end{equation}
and the intent $i_{t+1}$ may be revised based on the newly acquired information.

This process is repeated for multiple steps, enabling the agent to progressively refine its understanding of the document and uncover diverse and informative content. The explicit modeling of intent allows for flexible and open-ended exploration, permitting additional interactions when necessary.

The output of the exploration stage is a trajectory
\begin{equation}
\xi = \{(i_t, u_t, y_t)\}_{t=1}^{T},
\end{equation}
which serves as structured evidence for downstream QA generation.

In the exploration stage, each exploration step can be viewed as a random walk over the knowledge graph implicitly embedded in the entire document. 
When the number of such walks is sufficiently large, this process can, in principle, reconstruct the underlying document-level knowledge graph in a reverse manner. 
This idea is conceptually aligned with prior work on QA generation based on knowledge graphs in web search agent~\cite{li2025websailor,li2025websailorv2}. 
We do not explicitly construct a document-level knowledge graph in advance, as such an approach would incur substantial engineering complexity and overhead. 
Instead, our method adopts a more lightweight design that is nevertheless capable of generating challenging DocQA pairs, achieving a better trade-off between efficiency and effectiveness.

\textbf{\textit{Synthesis} Stage.}
Given the exploration trajectory $\xi$, the agent enters the synthesis stage to generate document-grounded QA pairs.
A synthesis model $M_s$ performs reasoning over the accumulated observations and generates a QA pair:
\begin{equation}
(q, a) \sim M_s(\xi, \mathcal{D}),
\end{equation}

This stage emphasizes \textit{(i)} reasoning over multiple observations collected during exploration, \textit{(ii)} grounding both questions and answers in the source document, and \textit{(iii)} producing semantically coherent and well-formed outputs.
The final output is a set of $K$, document-grounded QA pairs:
\begin{equation}
\mathcal{QA} = \{(q_k, a_k)\}_{k=1}^{K},
\end{equation}
which can be used for training an agent.
We employ a strong open-source model $M_t$ to perform rejection sampling over these QA pairs, $\mathcal{QA}$, thereby obtaining high-quality training trajectories.

\subsection{Agent Training}
Following the empirical findings of ~\cite{chen2023fireact}, twe mask loss contributions from observation tokens to mitigate interference from external feedback during training, which has been shown to improve both performance and robustness.
Given the task context \(\mathbf{tc}\) and the complete execution trajectory $\mathcal{H}=(x_0,...,x_{n-1},x_n)$, where each \(x_i \in \{ \tau, \alpha, o \} \), the loss $L$ is computed as follows:
\begin{equation}
\begin{aligned}
L = 
- \frac{1}{\sum_{i=1}^{|\mathcal{H}|} \mathbb{I}[x_i \ne o]}
\sum_{i=1}^{|\mathcal{H}|} 
\mathbb{I}[x_i \ne o] \cdot \\
\log \pi_{\theta}(x_i \mid \mathbf{tc}, x_{<i})
\end{aligned}
\end{equation}

Here, \(  \mathbb{I}[x_i \ne o] \) filters out tokens corresponding to external feedback, ensuring the loss is computed only over the agent’s decision steps.
\begin{table*}[t]
\centering
\small
\begin{tabular}{l|l|cccc}
\toprule
\multirow{2.05}{*}{\textbf{Method}} & \multirow{2.05}{*}{\textbf{Model}} & \multicolumn{3}{c}{\textbf{MMLongBench-Doc}} & \multicolumn{1}{c}{\textbf{DocBench}}  \\
&&\textit{acc} & \textit{$F_{1}$} & \textit{LasJ} &\textit{LasJ} \\
\midrule

\multicolumn{5}{l}{\textit{VLM Baseline}} \\
\arrayrulecolor{black!20}\midrule
 Naive VL ~\cite{ma2024mmlongbench} & \texttt{GPT-4o} & 42.8 & 44.9 & -- & 63.1 \\
 Naive VL~\cite{zhu2025doclens}  & \texttt{Gemini-2.5-Pro} &  -- & -- & 58.1 & -- \\
\arrayrulecolor{black}\midrule

\multicolumn{5}{l}{\textit{OCR-based Baseline}} \\
\arrayrulecolor{black!20}\midrule
fitz\footnote{\url{0https://pypi.org/project/fitz/}} & \texttt{GPT-4} & -- & -- & -- & 67.9 \\
Tesseract~\cite{smith2007overview} & \texttt{GPT-4o} & 30.1 & 30.5 & -- & -- \\
Tesseract~\cite{smith2007overview} & \texttt{Gemini-2.0-Flash} & 39.6 & 37.2 & -- & -- \\
\arrayrulecolor{black}\midrule
\multicolumn{5}{l}{\textit{RAG-based Baseline}} \\
\arrayrulecolor{black!20}\midrule
VisRAG~\cite{yu2024visrag} & \texttt{GPT-4o} & 29.0 & 27.8 & -- & --\\
Colpali~\cite{faysse2024colpali} & \texttt{GPT-4o} & 32.2 & 30.8 & -- & -- \\
M3DocRAG w/ ColPali~\cite{cho2025m3docvqa} & \texttt{Qwen2-VL-7B} & 31.4 & 36.5 & -- & -- \\
RAGAnything~\cite{guo2025rag} & \texttt{GPT-4o-mini} & 42.8 & -- & -- & 63.4\\
\arrayrulecolor{black}\midrule

\multicolumn{5}{l}{\textit{Prompt-based Agent}} \\
\arrayrulecolor{black!20}\midrule
Doc-React~\cite{wu2025doc} & \texttt{GPT-4o} & 38.1 & 38.3 & -- & -- \\
MDocAgent~\cite{han2025mdocagent} & \texttt{GPT-4o} &  42.0 & -- & -- & -- \\
MACT~\cite{yu2025visual} & \texttt{MiMo-VL-7B} & 47.4 & -- & -- & -- \\
SimpleDoc~\cite{jain2025simpledoc} & \texttt{Claude-4-Sonnet} & -- & -- & 58.6 & -- \\
SimpleDoc~\cite{jain2025simpledoc} & \texttt{Gemini-2.5-Pro} & -- & -- & 56.6 & -- \\
DocLens~\cite{zhu2025doclens} & \texttt{Claude-4-Sonnet} & -- & -- & 63.3 & -- \\
DocLens~\cite{zhu2025doclens} & \texttt{Gemini-2.5-Pro} &  -- & -- & \textbf{67.6} & -- \\
DocAgent~\cite{sun-etal-2025-docagent}
& \texttt{GPT-4o} & 51.8 & 49.1 & -- & 79.9\\
DocAgent~\cite{sun-etal-2025-docagent}
& \texttt{Claude-3.5-Sonnet} & \textbf{57.3} & 54.1 & -- & --  \\
\arrayrulecolor{black}\midrule
\multicolumn{5}{l}{\textit{Ours}} \\
\arrayrulecolor{black!20}
\midrule
\multirow{5}{*}{\textbf{DocDancer}} & \texttt{GPT-4o} & 52.3 & 50.8 & 59.2 & 73.5  \\
& \texttt{Gemini-2.5-Pro} & 56.3 & 55.3 & \underline{65.9} & 79.9 \\ 
 & \cellcolor{gray!10}\texttt{GPT-5.2} &  \cellcolor{gray!10}\underline{57.0} & \cellcolor{gray!10}\textbf{56.8}  & \cellcolor{gray!10}\textbf{67.6} & \cellcolor{gray!10}\textbf{85.5} \\
& \cellcolor{blue!10}\texttt{Qwen3-4B (ft)} & \cellcolor{blue!10}48.4 & \cellcolor{blue!10}49.2 & \cellcolor{blue!10}59.4 & \cellcolor{blue!10}79.8 \\
& \cellcolor{blue!20}\texttt{Qwen3-30B-A3B (ft)} & \cellcolor{blue!20}54.4 & \cellcolor{blue!20}53.9 & \cellcolor{blue!20}65.3 & \cellcolor{blue!20}\underline{81.2} \\
\arrayrulecolor{black}\midrule
Human Baseline  & --     &  65.8 & 66.0 & -- & \underline{81.2} \\
\bottomrule
\end{tabular}
\caption{\textbf{Performance comparison} across two long-context understanding benchmarks.
The best results among all methods are \textbf{bolded} and the second-best results are \underline{underlined}. 
}
\label{tab:docagent_results}
\end{table*}
\section{Experiments}
In this section, we aim to answer the following research questions (\textbf{RQs}):
\begin{itemize}[topsep=0pt,itemsep=0pt,parsep=0pt,leftmargin=*]
\item \textbf{RQ1}: How effective is the proposed information-seeking agent framework for DocQA?
\item \textbf{RQ2}: How effective is the proposed synthetic data pipeline for training open-source DocQA agents?
\item \textbf{RQ3}: Which components of the agent framework contribute most to performance?
\item \textbf{RQ4}: How does the proposed DocDancer in qualitative evaluations?
\end{itemize}

\subsection{Experimental Setup}
We fine-tune Qwen3-30B-A3B-Thinking-2507 and Qwen3-4B-Thinking-2507 on our dataset, resulting in DocDancer.
Our detailed implementation is provided in Appendix~\ref{app:imple}, trained with only 5,000 agent trajectories.

\noindent \textbf{Benchmarks.} We evaluate the proposed DocAgent on two multimodal long-context document question answering benchmarks: MMLongBenchDoc~\cite{ma2024mmlongbench} and DocBench~\cite{zou2025docbench}. MMLongBenchDoc comprises 135 documents with an average length of 47.5 pages, featuring rich layouts and multimodal components across seven diverse domains.
The dataset includes 1,091 questions derived from multiple sources, such as text, tables, charts, and images, with 33\% involving cross-page reasoning. DocBench consists of 229 real-world documents and 1,082 questions, covering five domains and four major question types.

\begin{figure*}[t]
  \centering
   \includegraphics[width=0.98\linewidth]{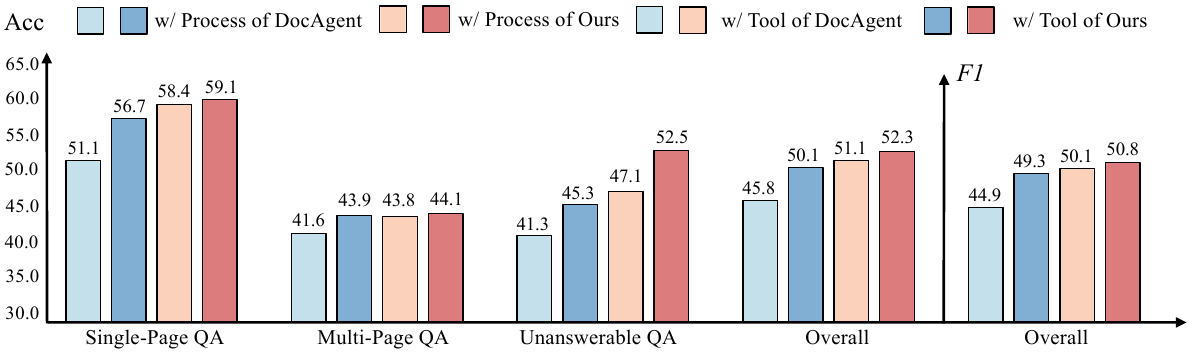}
   \caption{\textbf{Ablation study} on document parsing and tools.}
   \label{fig:Ablation_experiment}
\end{figure*}

\noindent \textbf{Metrics.} For MMLongBench-doc, we follow the official evaluation protocol. Answers are extracted using \texttt{GPT-4.1} and evaluated with rule-based scoring to compute F1 (\textit{$F_1$}) and Accuracy (\textit{acc}). To mitigate extraction errors and improve robustness to diverse response formats, we additionally employ an LLM-as-Judge (\textit{LasJ}) setting, where \texttt{gpt-4o} assigns binary scores using carefully designed prompts.
For DocBench, we likewise adhere to the official evaluation procedure, using the provided instructions to guide \texttt{GPT-4.1} for assessment.

\noindent \textbf{Baselines.} We compare our approach with the following three categories of baselines:
(1) VLM-based methods: Following the setting of MMLongBench-Doc, PDF pages are scanned at 144 DPI and used as input to the VLM.
(2) OCR-based methods: Text is extracted from documents using an OCR tool, and the parsed plain text is provided to a LLM for answering. Text beyond the model's context length is truncated.
(3) RAG-based methods: In this category, we compare existing RAG frameworks for DocQA, including VisRAG~\cite{yu2024visrag}, Colpali~\cite{faysse2024colpali}, M3DocRAG~\cite{cho2025m3docvqa}, MMGR~\cite{wan2025mmgraphrag}, and RAGAnything~\cite{guo2025rag}.
(4) Agent-based methods: We include several recent and well-performing training-free agentic frameworks, namely Doc-React~\cite{wu2025doc}, MDocAgent~\cite{han2025mdocagent}, MACT~\cite{yu2025visual}, SimpleDoc~\cite{jain2025simpledoc}, DocLens~\cite{zhu2025doclens}, and DocAgent~\cite{sun-etal-2025-docagent}.
The detailed introduction of the baseline is provided in Appendix~\ref{app:baselines}.

\subsection{Overall Performance (RQ1)}
 \begin{figure}[t]
  \centering
   \includegraphics[width=\linewidth]{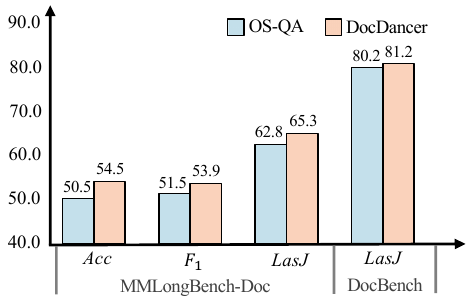}
   \caption{Performance comparison between models trained on \textbf{our synthesized QA data} and \textbf{open-source QA data.}}
\label{fig:baseline_vs_DocDancer_avg}
\end{figure}

\begin{figure}[t]
  \centering
\includegraphics[width=\linewidth]{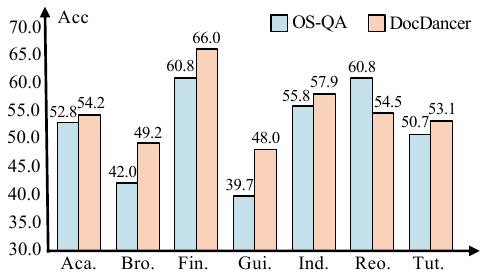}
   \caption{\textbf{Detailed domain-wise performance} comparison on MMLongBench-Doc between DocDancer and the model trained on OS-QA.}
   \label{fig:domain_performance}
\end{figure}

We evaluate our agent framework against OCR-based, RAG-based, and prompt-based baselines on long-document DocQA benchmarks.
Based on the experimental results in Table~\ref{tab:docagent_results}, we draw the following observations.
\textbf{First}, agent-based approaches substantially outperform VLM-based methods, OCR-based baselines, and RAG-based baselines across evaluated benchmarks, highlighting the advantage of explicit tool use and iterative reasoning for long-context document understanding.
\textbf{Second}, under the same backbone, our single-agent framework matches or surpasses multi-agent systems.
In particular, on MMLongBench-Doc, DocDancer with GPT-5.2 attains 56.8 $F_1$ / 67.6 \textit{LasJ}, outperforming all prior methods, and on DocBench, it reaches 85.5, exceeding the human baseline by 4 points.
\textbf{Third}, models trained on our synthetic DocQA dataset demonstrate strong generalization and data efficiency. Even with relatively small model sizes, such as 30B-A3B and 4B, the resulting agents achieve performance competitive with closed-source models. These results indicate that training agentic capabilities on smaller-scale models is both feasible and highly valuable, substantially lowering the barrier to building effective document-understanding agents.

\begin{figure*}[t]
  \centering
\includegraphics[width=\linewidth]{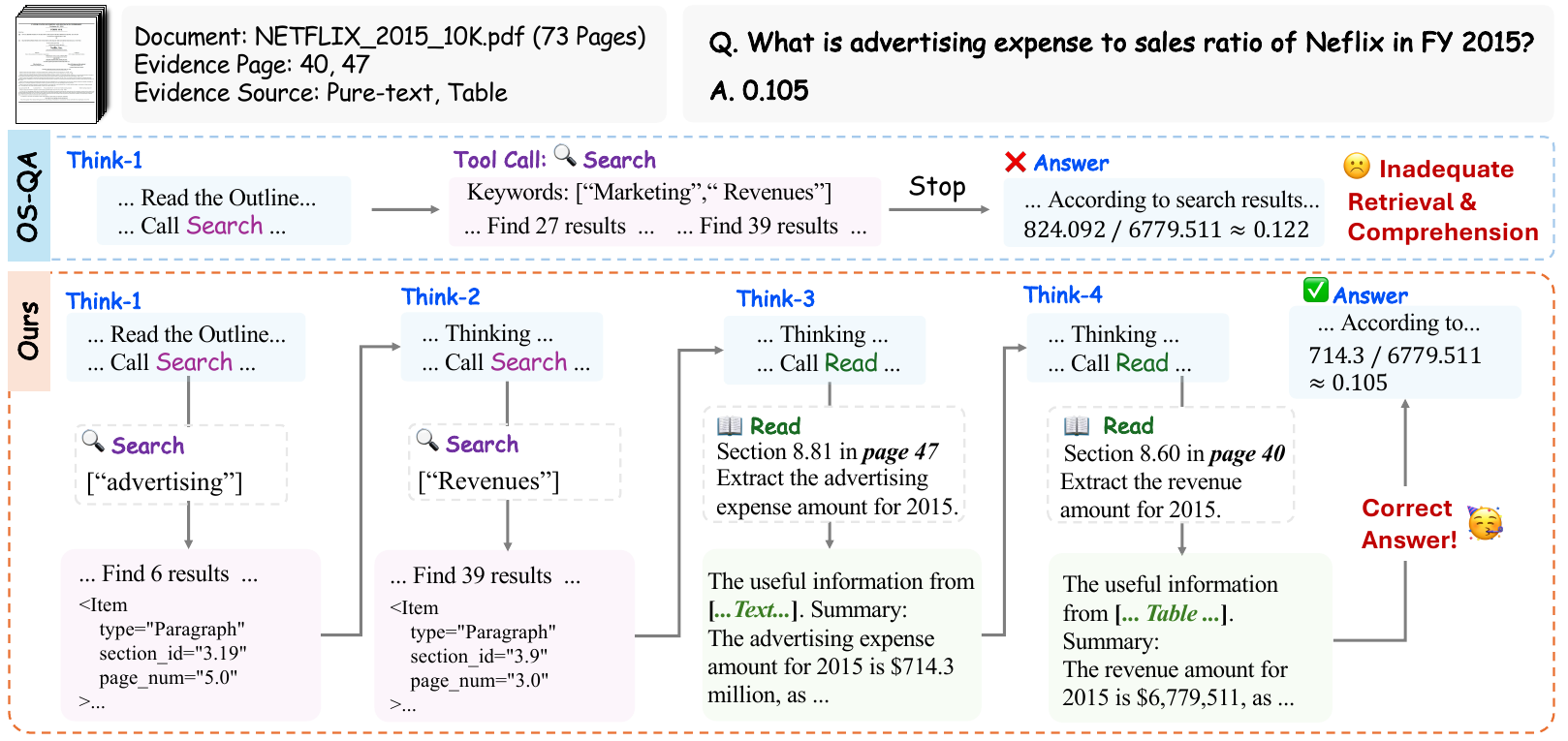}
   \caption{\textbf{A case study} demonstrating that our proposed DocDancer successfully performs multi-round information gathering to reach the correct answer, as illustrated in Table~\ref{app:detail} in detail, whereas OS-QA produces an incorrect result.}
   \label{fig:qa_case_study}
\end{figure*}

\subsection{Effectiveness of Synthetic Data (RQ2)}

\begin{figure}[t]
  \centering
\includegraphics[width=\linewidth]{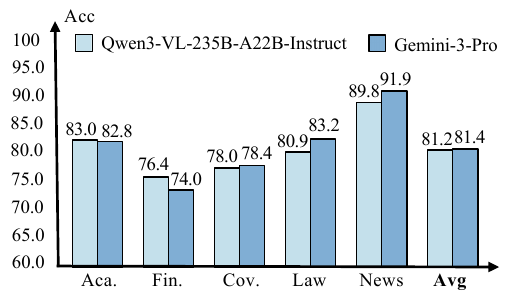}
   \caption{Results on DocBench across various domains \textbf{using different models used by \textit{Read} tool}. We report the generalized accuracy of five types of document domains, including Academia (Aca.), Finance (Fin.), Government (Gov), Law, and News.}
   \label{fig:tool_performance}
\end{figure}
\noindent \textbf{Overall Performance.} 
We investigate whether the \textit{Exploration-then-Synthesis} data generation pipeline provides effective supervision for learning agentic behaviors, and whether models trained solely on the synthesized data achieve strong performance compared to existing open-source QA pairs.
In Figure~\ref{fig:baseline_vs_DocDancer_avg}, we use the same PDF sources (Section \S\ref{sec:data}) and construct two training sets of \textbf{equal size} (5,000 instances): one from our synthesized QA data and the other from human-annotated QA data provided with the PDFs (\textbf{OS-QA}).
Both models are trained on Qwen3-30B-A3B-Thinking-2507.
Overall, DocDancer consistently outperforms OS-QA across all metrics and benchmarks, demonstrating the effectiveness of our data synthesis strategy.

\noindent \textbf{Detailed Results on Domains.}
Figure~\ref{fig:domain_performance} reports domain-level results on MMLongBench-Doc.
DocDancer consistently outperforms the QA baseline across all document domains, including Academic, Financial, Industry, and Report.
The gains are more pronounced in structurally complex domains that require iterative information seeking and fine-grained reasoning.
Overall, the results indicate that DocDancer generalizes well across diverse document types and is robust to domain variation.

\subsection{Influence of Agentic Tools (RQ3)}
\label{sec:ana_tool}
We conduct ablation studies on document processing for outline construction and tool usage in Figure~\ref{fig:Ablation_experiment}. 
The baseline is the Actor Agent from DocAgent~\cite{sun-etal-2025-docagent}. 
For outline construction, DocAgent relies on Adobe PDF Extract as well as DocXChain~\cite{yao2023docxchain} and PyMuPDF. 
In contrast, our enhanced method employs MinerU2.5~\cite{niu2025mineru2} for outline generation. 
The results demonstrate that, when combined with the same tools, our processing approach consistently outperforms the baseline, confirming that MinerU2.5 produces higher-quality document outlines.
Regarding tool usage, DocAgent utilizes five tools: \textit{search}, \textit{get\_section\_content}, \textit{get\_image}, \textit{get\_page\_images}, and \textit{get\_table\_image}. 
In comparison, we only use two tools, \textit{Search} and \textit{Read}, following the principle of simplicity. 
Despite this reduced tool set, our approach achieves better performance when combined with either our own outline or the outline generated by DocAgent.
The best results are obtained by combining our outline construction with our tool design, demonstrating their complementary effects.
Furthermore, we conduct an ablation study on the external model used by the \textit{Read} tool. 
Our default configuration, $M_m$ employs Qwen3-VL-235B-A22B-Instruct.
Replacing it with Gemini-3-Pro yields a modest overall improvement of 0.2 accuracy points on DocBench (Figure~\ref{fig:tool_performance}), with gains in Government, Law, and News domains.
These results indicate that our tool design is robust and does not depend on an exceptionally strong external model.

\subsection{Qualitative Analysis (RQ4)}
We present a case study of a financial task on a 73-page document from MMLongBench-Doc, as illustrated in Figure~\ref{fig:qa_case_study}.
Answering this question requires locating advertising expense and revenue figures from different sections of the document and performing a numerical computation.
The baseline model, which is trained on OS-QA relies on keyword-based retrieval and retrieves passages related to \emph{``marketing''} and \emph{``revenues''}. 
Due to insufficient grounding, it incorrectly uses a marketing expense figure as a proxy for advertising expense, yielding an erroneous ratio of $0.122$. 
This failure illustrates the limitation of single-pass retrieval and shallow aggregation when fine-grained financial concepts are required.
In contrast, DocDancer performs multi-round, question-driven information gathering. 
It first retrieves and reads the section explicitly reporting advertising expense for FY~2015 (\$714.3M), and then independently extracts the total revenue from a separate tabular section (\$6{,}779.5M). 
By grounding each value to its corresponding evidence and verifying semantic relevance, the system computes the correct ratio of $714.3 / 6{,}779.5 \approx 0.105$.
It demonstrates that accurate document-level financial question answering benefits from our synthetic data, which enables the construction of \textbf{domain-specific expert-level} supervision beyond ordinary human annotations.
\section{Conclusion}
We propose DocDancer, an end-to-end trained agentic model for document question answering that formulates DocQA as an information-seeking process.
By introducing a tool-centric framework with complementary search and read operations, DocDancer enables effective exploration and comprehension of long, structured documents.
To mitigate the lack of high-quality supervision, we further design an Exploration-then-Synthesis data pipeline that generates compact yet effective training data for learning agentic behaviors.
Experiments on MMLongBench-Doc and DocBench demonstrate that DocDancer achieves strong and competitive performance, validating the effectiveness of agentic information-seeking for document understanding.

\section*{Limitations}
This work still has several limitations.
First, our experiments are conducted only on Qwen3-30B-A3B-Thinking-2507 and Qwen3-4B-Thinking-2507; we do not evaluate the proposed method on larger-scale models or models from other families.
Second, we focus exclusively on supervised fine-tuning (SFT) and do not explore agentic reinforcement learning (RL).
Third, we do not further scale the training data, and thus do not investigate how the proposed method performs under larger or more diverse data.

\section*{Ethical Considerations}
This work studies agentic document-grounded question answering using publicly available benchmarks and documents released for research purposes. 
The proposed \textit{Exploration-then-Synthesis} pipeline generates synthetic question–answer pairs that are explicitly grounded in source documents and does not introduce new proprietary data or attempt to reproduce large portions of copyrighted text verbatim.
While the method itself does not collect personal information, document-grounded agents may be applied to sensitive or private documents in downstream use; such applications require appropriate authorization and privacy safeguards. 
The synthesized data and trained models may inherit biases present in the underlying document sources, including domain and content imbalances. 
Finally, although improved document exploration capabilities could be misused if deployed irresponsibly, the strong grounding in retrieved evidence and our commitment to releasing code and data aim to support transparency, reproducibility, and responsible research use.

\bibliography{custom}

\appendix

\clearpage
\newpage

\section{Case Study of Synthetic Data}
\label{app:case_study}
\begin{figure*}[t]
  \centering
\includegraphics[width=\linewidth]{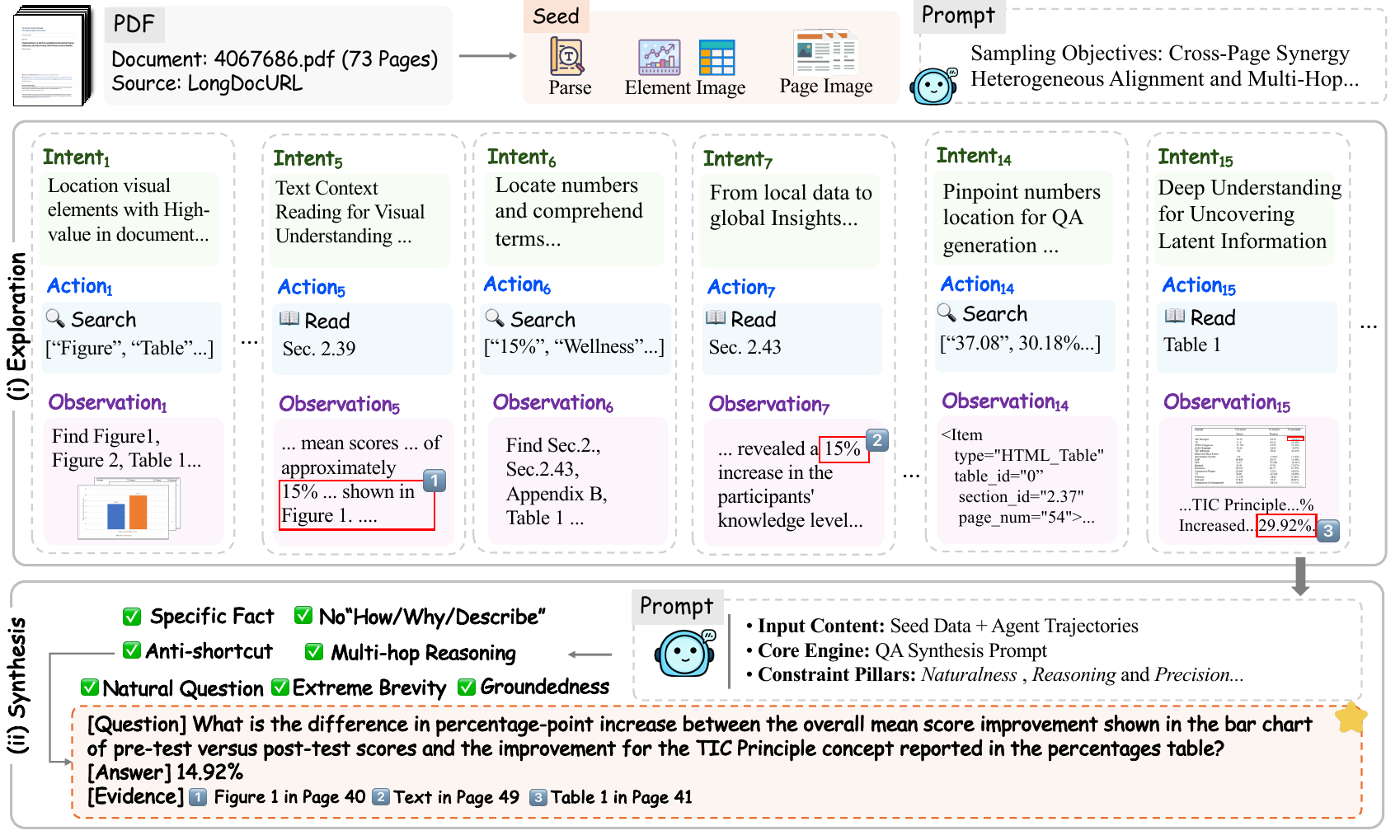}
   \caption{\textbf{A case study} of the Exploration-then-Synthesis framework generating a multi-hop, cross-document, cross-modal numerical reasoning QA pair.
}
   \label{fig:syn_case_study}
\end{figure*}

Figure~\ref{fig:syn_case_study} demonstrates how the \textit{Exploration-then-Synthesis} framework iteratively navigates a 73-page document, aggregating heterogeneous evidence, \underline{text} (in Sec. 2.43), \underline{charts} (in Figure 1), and \underline{tables} (in Table 1), scattered across disjoint pages (pp. 40, 41, 49) to synthesize a high-quality question that requires complex reasoning.

In the \textit{Exploration} Stage, the agent generates a exploartion trajectory $\xi$ via iterative $(i_t, u_t)$ steps, effectively performing a ``random walk'' over the document's implicit knowledge graph.
It aggregates heterogeneous evidence by bridging disjoint pages—linking visual trends in a chart (p.~40) with precise values in text (p.~49) and a table (p.~41). 
In the \textit{Synthesis} Stage, the model $M_s$ reasons over this accumulated trajectory to construct a complex multi-hop numerical question~\cite{wu-etal-2025-webwalker}.
The final QA pair requires arithmetic calculation ($29.92\% - 15\% = 14.92\%$) rather than simple retrieval, ensuring deep document grounding and preventing shortcut learning.

\section{Implementation Details}
\label{app:imple}

\subsection{Details on Prompts}
The prompts for the DocDancer are shown in Figure~\ref{fig:agent_system_prompt}.

\definecolor{PromptFrame}{HTML}{44935b}  
\definecolor{PromptBack}{HTML}{ddeed9}   
\begin{figure}[!htp]
    \centering
\begin{tcolorbox}[
    breakable,
    title=Prompt,
    colback=PromptBack!20,   
    colframe=PromptFrame, 
    fonttitle=\bfseries,
]
You are an expert research assistant tasked with answering questions based on document content. \\
You will be provided with an XML outline of the document.
If you need more comprehensive, detailed, or accurate information from the document to fully address the user's query, you need to use the provided tool.
\\
\\
I've uploaded a document, and below is the outline in XML format:
\{document\_outline\}.\\
Answer the following question based on the content of the document:
\{question\}.
\end{tcolorbox}
    \caption{System prompt for \textbf{DocDancer}.}
    \label{fig:agent_system_prompt}
\end{figure}

\subsection{Tool Schema}
This section details the tool schemas provided to the agent. We designed two primary tools: \texttt{search} for keyword-based retrieval and \texttt{read} for extracting content from specific document sections. The specific JSON structures defining these functions are shown in Figure~\ref{fig:tool_schema}.

\begin{figure*}[t] 
    \definecolor{solarizedBase}{RGB}{242, 245, 250} 
    \definecolor{solarizedFrame}{RGB}{081, 132, 178}  
    \definecolor{codeGreen}{RGB}{133, 153, 0}
    \definecolor{codeCyan}{RGB}{42, 161, 152}

    \lstset{
        basicstyle=\ttfamily\footnotesize,
        breaklines=true,
        columns=fullflexible,
        stringstyle=\color{codeCyan},
        keywordstyle=\color{codeGreen},
        upquote=true,
    }

    \begin{tcolorbox}[
        title={Tool Schemas},
        colback=solarizedBase!20,      
        colframe=solarizedFrame,    
        coltitle=white,
        fonttitle=\bfseries,
        width=\textwidth 
    ]
    \textbf{\textit{Search}}
    \begin{lstlisting}
{
    "type": "function",
    "function": {
        "name": "search",
        "description": "Find and extract all paragraphs and sections where any of the provided search terms appear",
        "parameters": {
            "type": "object",
            "properties": {
                "keywords": {
                    "type": "array",
                    "items": {
                        "type": "string"
                    },
                    "description": "A list of query keywords for searching"
                }
            },
            "required": ["keywords"]
        }
    }
}
\end{lstlisting}
\textbf{\textit{Read}}
\begin{lstlisting}
{
    "type": "function",
    "function": {
        "name": "read",
        "description": "Read multiple sections by section IDs and extract useful information from all content contained in those sections, including both visual elements and textual elements.",
        "parameters": {
            "type": "object",
            "properties": {
                "section_ids": {
                    "type": "array",
                    "items": {
                        "type": "string"
                    },
                    "description": "A list of section IDs to read from the document"
                },
                "goal": {
                    "type": "string",
                    "description": "The user goal that guides what useful information should be extracted from the selected sections"
                }
            },
            "required": ["section_ids", "goal"]
        }
    }
}
    \end{lstlisting}
    \end{tcolorbox}
    \caption{Tool schema: \textit{Search} and \textit{Read}.}
    \label{fig:tool_schema}
\end{figure*}

\subsection{Training Details}
We fine-tune Qwen3-30B-A3B-Think\footnote{https://huggingface.co/Qwen/Qwen3-30B-A3B-Thinking-2507} and Qwen3-4B-Think\footnote{ https://huggingface.co/Qwen/Qwen3-4B-Thinking-2507} using the Megatron-LM framework~\cite{zhao2024swiftascalablelightweightinfrastructure,megatron-lm}.
Both models are trained with a context length of 128k to support long-document processing tasks. 
We employ the AdamW optimizer with a precision-aware configuration and a cosine decay learning rate scheduler, featuring a peak learning rate of $1.0 \times 10^{-5}$, a minimum of $1.0 \times 10^{-6}$, and a 5\% warmup phase. 
The global batch size is configured to 16 for the Qwen3-30B-A3B-Think and to 40 for Qwen3-4B-Think. 
For Qwen3-30B-A3B-Think, we apply an auxiliary loss coefficient of $10^{-3}$ to ensure balanced expert routing.
We train both models for 10 epochs and selected the checkpoint with best performance.

\subsection{Inference Details}
\textit{vLLM} framework~\cite{kwon2023efficient} is used for
inference; we employ a temperature of 0.6, a $top_p$ value of 0.95, and a presence penalty of 1.1.
\subsection{Hyperparameter}
By default, $M_m$ is Qwen3-VL-235B-A22B-Instruct, and we analyze the effects of replacing it in Section~\ref{sec:ana_tool}.
For $M_t$, we use the open-source and relatively strong model gpt-oss-120b to perform rejection sampling.
Further analysis is provided in Table~\ref{tab:ana_oss}.
First, our method substantially outperforms the base model without fine-tuning, demonstrating the effectiveness of the proposed training strategy.
Second, our approach also surpasses the model trained with reject sampling, validating the quality of the synthesized question–answer data and showing that it can effectively elicit and enhance the model’s performance.
For $M_s$, we employ gpt-oss-120b in \textit{Exploration-then-Synthesis} framework to synthesis data.

\subsection{Details on Prompts for Data Synthesis}

The prompts utilized for \textbf{Exploration} and \textbf{Synthetic} within the Exploration-then-Refine framework are presented in Figure \ref{fig:syn_prompt_traj} and Figure \ref{fig:syn_prompt_qa}, respectively.
Regarding the exploration configuration, we adjust the maximum exploration depth based on the complexity of the document sources. Specifically, we set the maximum sampling depth to 20 for LongDocURL and MMdocRAG, while for DUDE and CUAD, this limit is set to 15.

\begin{table*}[t]
\centering
\small
\begin{tabular}{l|l|cccc}
\toprule
\multirow{2.05}{*}{\textbf{Method}} & \multirow{2.05}{*}{\textbf{Model}} & \multicolumn{3}{c}{\textbf{MMLongBench-Doc}} & \multicolumn{1}{c}{\textbf{DocBench}}  \\
&&\textit{acc} & \textit{$F_{1}$} & \textit{LasJ} &\textit{LasJ} \\
\midrule
DocDancer & \texttt{Qwen3-A3B-30B-Thinking} & 39.2 & 36.4 &  46.9 & 74.1 \\
DocDancer & \texttt{GPT-oss-120B} & 52.3 & 53.0 & 59.8  & 80.8 \\
DocDancer & \cellcolor{blue!20}\texttt{Qwen3-30B-A3B-Thinking (ft)} & \cellcolor{blue!20}54.4 & \cellcolor{blue!20}53.9 & \cellcolor{blue!20}65.3 & \cellcolor{blue!20}81.2 \\
\bottomrule
\end{tabular}
\caption{\textbf{Performance comparison} across two long-context understanding benchmarks.}
\label{tab:ana_oss}
\end{table*}

\begin{figure*}[ht] 
    \centering
    
    \definecolor{TrajFrame}{HTML}{734C81} 
    \definecolor{TrajBack}{HTML}{F4ECF4}   

    \begin{tcolorbox}[
        title=\textit{Exploration} in Exploration-then-Refine Framework., 
        fontupper=\small,
        colback=TrajBack!20,    
        colframe=TrajFrame!80,  
        coltitle=white,      
    ]
        You are exploring a parsed PDF paper/report (outline + paragraphs + images + table snapshots + per-page screenshots). Your objective is to collect HIGH-QUALITY, GROUNDED evidence bundles that can later support HARD, multi-hop, visually grounded document Q\&A synthesis.

        \medskip
        \textbf{Final QA Constraints You Must Enable (every eventual QA must satisfy ALL):}
        \begin{itemize}[leftmargin=*, nosep]
            \item Multi-page: Combining evidence from at least THREE different pages/sections, where the pieces of evidence are related.
            \item Multi-element: Contains at least two evidence source types (text paragraphs/charts/graphics/table screenshots and/or full-page layouts).
            \item Multi-hop: require at least TWO reasoning points (e.g. cross-reference + computation, footnote rule + chart reading, layout count + comparison, multiple related searches + readings).
        \end{itemize}

        \medskip
        \textbf{Important:} final questions should NOT rely on explicit document locations.
        Do NOT plan to use page numbers, section titles/IDs, or explicit figure/table numbers (e.g., ``Figure $<$number$>$'', ``Table $<$number$>$'') in the question.
        Instead, you must collect CONTENT-BASED CLUES that can uniquely identify the needed evidence:
        \begin{itemize}[leftmargin=*, nosep]
            \item Caption keywords (short quote fragments), axis labels and units, legend item names, panel labels (a)/(b), distinctive row/column headers, and footnote phrases (``restated'', ``excluding'', ``unaudited'', unit changes).
        \end{itemize}

        \medskip
        \textbf{Exploration strategy using only search and read:}
        \begin{itemize}[leftmargin=*, nosep]
            \item Use search to find visuals, tables, footnotes, and their nearby discussion text.
            Start with keywords like: ``Figure'', ``Fig.'', ``Chart'', ``Image'', ``Graph'', ``legend'', ``axis'', ``panel'', ``Table'', ``Note'', ``footnote'', ``restated'', ``excluding'', ``unaudited''.
            \item For each promising hit, immediately read the covering section(s) with a goal that extracts:
            \begin{itemize}[leftmargin=1em, nosep]
                \item The text content of the section in question.
                \item Caption text, axis labels/units, legend items, and visual markers.
                \item The exact table header path, target cell(s), and footnote rules.
                \item The narrative claim/explanation that references the visual.
            \end{itemize}
            \item Use the read function as much as possible, deliberately chain across pages.
            \item For conditional layout questions: identify a page by a unique visual cue, then use read to count visible tables/figures.
        \end{itemize}

        \medskip
        \textbf{Avoid:}
        \begin{itemize}[leftmargin=*, nosep]
            \item Broad whole-document counts unless you turn them into comparative, multi-hop questions.
            \item Word-frequency counting.
            \item Repeating identical tool calls.
            \item Statistical analysis of the number of elements.
        \end{itemize}

        Every action during sampling should contribute to forming a future HARD, multi-page, multi-element, multi-hop document QA.
    \end{tcolorbox}
    \caption{Prompt for \textit{exploration} stage in Exploration-then-Refine framework.}
    \label{fig:syn_prompt_traj}
\end{figure*}

\begin{figure*}[ht]
    \centering
    
    \definecolor{TrajFrame}{HTML}{734C81} 
    \definecolor{TrajBack}{HTML}{F4ECF4}   

    \begin{tcolorbox}[
        title=\textit{Synthesis} in Exploration-then-Refine Framework., 
        fontupper=\small,
        colback=TrajBack!20,    
        colframe=TrajFrame!80,  
        coltitle=white,      
    ]
        You must synthesize ``document Q\&A'' training data based ONLY on the trajectory.

        \medskip
        \textbf{Hard Requirements (Strict):}
        \begin{itemize}[leftmargin=*, nosep]
            \item The output must be a JSON object containing only two fields: \texttt{question} and \texttt{answer} (no additional fields are allowed), and must be in English only.
            \item The question must be natural and unambiguous, containing only one question and corresponding to a single, unique answer.
            \item The question must not be a common-knowledge question; it must be impossible to answer based on the question alone and must be highly dependent on the document.
            \item Do not mention tools, sections, pages, section IDs, searching/reading actions, trajectories, or observations.
            \item The answer length should be limited to a single sentence, ideally a short phrase, entity, number, or list, and avoid simply using ``yes/no'' answers. The answer must be directly supported by evidence from the provided text and cannot be guessed randomly.
        \end{itemize}

        \medskip
        \textbf{Mandatory Difficulty Constraints (every QA pair must satisfy all of the following):}
        \begin{enumerate}[leftmargin=*, nosep]
            \item \textbf{Multi-page:} The question requires evidence from at least two different pages/sections to answer, and the evidence must be logically related.
            \item \textbf{Multiple Evidence Modalities:} The question must involve at least two types of evidence, such as text, charts, figures, tables, screenshots, and/or full-page layout cues, with a preference for covering visual elements.
            \item \textbf{Multi-step Reasoning:} The question must require at least two reasoning steps (e.g., calculation + cross-validation, footnote rule application + chart reading, layout counting + comparison).
        \end{enumerate}

        \medskip
        \textbf{No Explicit Location References in the Question:}
        \begin{itemize}[leftmargin=*, nosep]
            \item Do not mention page numbers, section IDs, titles/IDs, or explicit figure/table numbers (e.g., ``Figure $<$number$>$'', ``Table $<$number$>$'').
            \item Instead, provide 1--3 content-based clues to help locate the evidence, such as: short title phrases, axis labels/units, legend item names, unique row names, footnote keywords, or distinctive layout hints (e.g., ``the only multi-panel figure labeled (a) and (b)'').
            \item When describing visual elements, do not directly copy long unique numbers or OCR-extracted long text strings from images (e.g., ``an image showing the number 7,584,322,338''). Use specific entity names or semantic descriptions instead (e.g., ``Apple's 2018 total sales table'', ``an image showing adjusted outstanding balances'', or ``the largest segment in the pie chart'').
        \end{itemize}

        \medskip
        \textbf{Preferred Question Templates (all templates must be cross-page + visual + multi-step):}
        \begin{itemize}[leftmargin=*, nosep]
            \item \textbf{Cross-page conditional layout:} Identify pages via unique visual cues and compare the number of visible objects across pages.
            \item \textbf{Textual claim + chart verification:} A narrative statement about a change/target that is verified using a chart and light calculation.
            \item \textbf{Table + chart consistency:} Compute a ratio/difference from a table and verify it against a data point in a chart on another page.
            \item \textbf{Footnote-constrained table + chart mapping:} Apply footnote/restatement/exclusion rules, then map the correct year/value to a chart on another page.
            \item \textbf{Table/Chart comprehension questions:} Locate tables and charts via text, then derive conclusions from table structure or chart visuals.
            \item \textbf{Unanswerable questions:} Questions that seem reasonable but are actually impossible to answer (e.g., questions about terms/entities that do not exist in the document). For these, the answer must be ``Unanswerable''.
            \item \textbf{Counting questions:} Count the occurrences of key local terms or entities in the document. Such questions should only be generated when there is sufficient and conclusive evidence.
        \end{itemize}

        \medskip
        \textbf{Fallback Rule:}
        \begin{itemize}[leftmargin=*, nosep]
            \item If the current trajectory cannot support a question that satisfies all constraints, choose a different question.
        \end{itemize}

        \medskip
        \textbf{After generating a question, perform a second-pass check and regenerate if the question falls into any of the following categories:}
        \begin{itemize}[leftmargin=*, nosep]
            \item Contains more than one question.
            \item Includes non-English languages or characters.
            \item Questions that can be answered based on an independent page/section.
            \item Common-sense questions unrelated to the document.
            \item Counting tasks spanning the entire document with a broad scope.
            \item Counting tasks involving Charts/Figures/Images/Tables.
        \end{itemize}

        \medskip
        For unanswerable questions, confirm that they are truly unanswerable.
        For counting questions, confirm completeness and answer accuracy.
        \textbf{Do not guess or fabricate answers under any circumstances.}
    \end{tcolorbox}
    \caption{Prompts for Q\&A \textit{Synthesis} stage in Exploration-then-Refine framework.}
    \label{fig:syn_prompt_qa}
\end{figure*}

\section{Baselines}
\label{app:baselines}

We compare DocDancer against a comprehensive set of baselines categorized into four groups:

\noindent \textbf{\textit{Naive VLM Baselines.}}
These methods evaluate the native long-context understanding capabilities of advanced VLMs. We directly feed PDF pages converted to images (144 DPI) into the models without external parsing or retrieval. 
Following the settings in MMLongBench-Doc~\cite{ma2024mmlongbench}, we report \textit{GPT-4o}~\footnote{\url{https://platform.openai.com/docs/models/gpt-4o}} and \textit{Gemini-2.5-Pro}~\footnote{\url{https://ai.google.dev/gemini-api/docs/models?\#gemini-2.5-pro}}.

\noindent \textbf{\textit{OCR-based Baselines.}}
These baselines treat the task as text-only QA by first extracting content using OCR engines. We pair \textit{Tesseract}~\cite{smith2007overview} and \textit{PyMuPDF (fitz)}~\footnote{\url{https://pymupdf.readthedocs.io/}} with LLMs including \textit{GPT-4}, \textit{GPT-4o}, and \textit{Gemini-2.0-Flash}.

\noindent \textbf{\textit{RAG-based Baselines.}}
We consider both visual and hybrid retrieval strategies:
\begin{itemize}
    \item \textbf{Visual Retrieval:} \textbf{VisRAG}~\cite{yu2024visrag} and \textbf{ColPali}~\cite{faysse2024colpali} retrieve relevant page or patch-level visual evidence based on vision-centric embeddings, utilizing \textit{GPT-4o} for response generation.
    \item \textbf{Hybrid Retrieval:} \textbf{M3DocRAG}~\cite{cho2025m3docvqa} performs joint retrieval using a multimodal retriever with \textit{Qwen2-VL-7B}. \textbf{RAGAnything}~\cite{guo2025rag} structures multimodal content as knowledge entities for cross-modal retrieval, using \textit{GPT-4o-mini} as the backbone.
\end{itemize}

\noindent \textbf{\textit{Prompt-based Agentic Baselines.}}
We include state-of-the-art agent frameworks designed for document understanding:
\begin{itemize}
    \item \textbf{Doc-React}~\cite{wu2025doc} employs an iterative decision-making process to balance information gain and uncertainty reduction (\textit{GPT-4o}).
    \item \textbf{MDocAgent}~\cite{han2025mdocagent} utilizes a multi-agent system with five specialized roles for context retrieval (\textit{GPT-4o}).
    \item \textbf{MACT}~\cite{yu2025visual} introduces a multi-agent collaboration framework featuring adaptive test-time scaling (\textit{MiMo-VL-7B}~\cite{team2025kimi}).
    \item \textbf{SimpleDoc}~\cite{jain2025simpledoc} retrieves pages via \textit{ColQwen2.5}, followed by LLM-based evidence selection (\textit{Claude-4-Sonnet}, \textit{Gemini-2.5-Pro}).
    \item \textbf{DocLens}~\cite{zhu2025doclens} operates as a tool-augmented multi-agent framework for focused reading (\textit{Claude-4-Sonnet}, \textit{Gemini-2.5-Pro}).
    \item \textbf{DocAgent}~\cite{sun-etal-2025-docagent} leverages a tree-structured document outline combined with retrieval tools (\textit{GPT-4o}, \textit{Claude-3.5-Sonnet}).
\end{itemize}

\onecolumn

\begin{longtable}{>{\raggedright\arraybackslash}p{0.95\linewidth}}
    
    \toprule
    \cellcolor{blue!20}{\textbf{Details of Case Study}} \\ 
    \midrule
    \endfirsthead

    \multicolumn{1}{c}{\small\textit{Continued from previous page}} \\
    \toprule
    \cellcolor{blue!20}{\textbf{Details of Case Study (Continued)}} \\ 
    \midrule
    \endhead

    \midrule
    \multicolumn{1}{r}{\small\textit{Continued on next page...}} \\
    \endfoot

    \bottomrule
    \noalign{\vspace{2em}}
    \caption{Case study of a financial task.} \label{app:detail} \\
    \endlastfoot

    \textbf{Question} \\
    \quad What is advertising expense to sales ratio of Netflix in FY 2015? Round your answer to three decimal places.\\
    \textbf{Answer}\\
    \quad 0.105 \\
    
    \midrule
    \multicolumn{1}{c}{\cellcolor{blue!10}\textit{Step 1}} \\
    \midrule
    
    \textcolor{codegreen}{\texttt{<think>}} \\
    \quad We need to answer: ``what is advertising expense to sales ratio of Netflix in FY 2015? Round your answer to three decimal places.'' We need to locate advertising expense and sales (revenues) for FY 2015. From outline we have various sections. First, find advertising expense. The term ``advertising'' appears in sections. Search for ``advertising''. \\
    \textcolor{codegreen}{\texttt{</think>}} \\
    
    \textcolor{codeblue}{\texttt{<tool\_call>}} \\
    \quad \texttt{\{ "arguments": "\{ "keywords": ["advertising"] \}", "name": "search" \}} \\
    \textcolor{codeblue}{\texttt{</tool\_call>}} \\
    
    \textcolor{codered}{\texttt{<tool\_response>}} \\
    \quad A Document search for `advertising` found 6 results: \\
    \quad \texttt{<Item type="Paragraph" section\_id="3.19" page\_num="5.0">}... adjust pricing or service offerings ... \texttt{</Item>}\\
    \quad ...\\
    \quad \texttt{<Item type="Paragraph" section\_id="8.81" page\_num="47.0">} \\
    \quad Marketing expenses consist primarily of advertising expenses... Advertising expenses were \$714.3 million, \$533.1 million and \$404.0 million for the years ended December 31, 2015, 2014 and 2013, respectively.\\
    \quad \texttt{</Item>} \\
    \textcolor{codered}{\texttt{</tool\_response>}} \\
    
    \midrule
    \multicolumn{1}{c}{\cellcolor{blue!10}\textit{Step 2}} \\
    \midrule
    
    \textcolor{codegreen}{\texttt{<think>}} \\
    \quad One of the results is in section\_id ``8.81'' paragraph page\_num ``47.0''. It says: ``Advertising expenses were \$714.3 million, \$533.1 million and \$404.0 million for the years ended December 31, 2015, 2014 and 2013, respectively.'' Thus advertising expense FY 2015 = \$714.3 million.\\
    \quad Now sales (revenues) for FY 2015: In consolidated statements of operations (section 8.60) we have revenues \$6,779,511 (in thousands). That's \$6,779,511 thousand = \$6,779.511 million.\\
    \quad Alternatively, there may be a line item ``Revenues'' in Table 5 etc. Let's locate ``Revenues''. \\
    \textcolor{codegreen}{\texttt{</think>}} \\
    
    \textcolor{codeblue}{\texttt{<tool\_call>}} \\
    \quad \texttt{\{ "arguments": "\{ "keywords": ["Revenues"] \}", "name": "search" \}} \\
    \textcolor{codeblue}{\texttt{</tool\_call>}} \\
    
    \textcolor{codered}{\texttt{<tool\_response>}} \\
    \quad A Document search for `Revenues` found 39 results:\\
    \quad \texttt{<Item type="Paragraph" section\_id="3.9" page\_num="3.0">}\\
    \quad ...\\
    \quad \texttt{<Item type="HTML\_Table" table\_id="5" section\_id="8.20" page\_num="19.0">} ...\\
    \quad Revenues (2015): \$ 6,779,511 ...\\
    \quad \texttt{</Item>}\\
    \quad ...\\
    \textcolor{codered}{\texttt{</tool\_response>}} \\

    \multicolumn{1}{c}{\cellcolor{blue!10}\textit{Step 3}} \\
    \midrule
    
    \textcolor{codegreen}{\texttt{<think>}} \\
    \quad We have advertising expense: \$714.3 million (2015). We need to compute ratio = advertising expense / revenue. Revenue for 2015: \$6,779,511 thousand = \$6,779.511 million. Thus ratio $= 714.3 / 6,779.511 = ?$ ... First, we should cite sources: advertising expense paragraph (8.81) and revenue table. \\
    Let's extract the advertising expense paragraph text to be sure. \\
    \textcolor{codegreen}{\texttt{</think>}} \\
    
    \textcolor{codeblue}{\texttt{<tool\_call>}} \\
    \quad \texttt{\{ "arguments": "\{ "section\_ids": ["8.81"], "goal": "Extract the advertising expense amount for 2015" \}", "name": "read" \}} \\
    \textcolor{codeblue}{\texttt{</tool\_call>}} \\
    
    \textcolor{codered}{\texttt{<tool\_response>}} \\
    \quad The useful information from the document section (section\_id=8.81) for user goal `Extract the advertising expense amount for 2015` is as follows:\\
    \quad Evidence in document:\\
    \quad Marketing expenses consist primarily of advertising expenses... Advertising expenses were \$714.3 million, \$533.1 million and \$404.0 million for the years ended December 31, 2015, 2014 and 2013, respectively.\\
    \quad Summary:\\
    \quad The advertising expense amount for 2015 is \$714.3 million... \\
    \textcolor{codered}{\texttt{</tool\_response>}} \\
    
    \midrule
    \multicolumn{1}{c}{\cellcolor{blue!10}\textit{Step 4}} \\
    \midrule
    
    \textcolor{codegreen}{\texttt{<think>}} \\
    \quad I have successfully extracted and verified the advertising expense (\$714.3 million) from section 8.81. Now, I need to formally verify the total revenue for 2015 to serve as the denominator. The search results in Step 2 indicated that section 8.60 contains the ``Consolidated Statements of Operations,'' which is the authoritative source for revenue figures.
    I will read section 8.60 to extract the exact revenue amount for the year ended December 31, 2015. \\
    \textcolor{codegreen}{\texttt{</think>}} \\
    
    \textcolor{codeblue}{\texttt{<tool\_call>}} \\
    \quad \texttt{\{ "arguments": "\{ "section\_ids": ["8.60"], "goal": "Extract the revenue amount for 2015" \}", "name": "read" \}} \\
    \textcolor{codeblue}{\texttt{</tool\_call>}} \\
    
    \textcolor{codered}{\texttt{<tool\_response>}} \\
    \quad The useful information from the document section (section\_id=8.60) for user goal `Extract the revenue amount for 2015` is as follows:\\
    \quad Evidence in document:\\
    \quad Heading: NETFLIX, INC.\\
    \quad Paragraph: CONSOLIDATED STATEMENTS OF OPERATIONS\\
    \quad \texttt{<HTML\_Table>} \\
    \quad \texttt{<table> ... <tr> <td>Revenues</td> <td>\$ 6,779,511</td> <td>\$ 5,504,656</td> ... </tr> ... </table>}\\
    \quad \texttt{</HTML\_Table>}\\
    \quad Caption: (in thousands, except per share data)\\
    \quad Summary: \\
    \quad The revenue amount for 2015 is \$6,779,511, as explicitly stated in the `Revenues' row of the `CONSOLIDATED STATEMENTS OF OPERATIONS' table for the year ended December 31, 2015. This value is presented in thousands, as indicated by the caption. \\
    \textcolor{codered}{\texttt{</tool\_response>}} \\
    
    \midrule    
    \multicolumn{1}{c}{\cellcolor{blue!10}\textit{Final Response}} \\
    \midrule   
    
    \quad Netflix’s advertising expense for fiscal 2015 was \textbf{\$714.3 million} [8.81], and its total revenue for the same year was \textbf{\$6,779,511 thousand} (i.e., \textbf{\$6,779.511 million}) [8.60]. \\
    \quad $\text{Advertising-to-sales ratio} = \frac{714.3}{6,779.511} \approx 0.10536$. \\
    \quad Rounded to three decimal places, the ratio is \textbf{0.105} (or 10.5\%). \\
    
\end{longtable}

\twocolumn

\end{document}